# Chapter : Towards technological adaptation of advanced farming through AI, IoT, and Robotics: A Comprehensive overview

Author(s):
Md. Mahadi Hasan (Undergraduate Researcher, Asian University of Bangladesh, mahadihasan@aub.edu.bd)
Muhammad Usama Islam (Lecturer, Asian University of Bangladesh, usamaislam@iut-dhaka.edu)
Muhammad Jafar Sadeq (Associate Professor, Asian University of Bangladesh, jafar@aub.edu.bd)

## Abstract

The population explosion of the 21st century has adversely affected the natural resources with restricted availability of cultivable land, increased average temperatures due to global warming, and carbon footprint resulting in a drastic increase in floods as well as droughts thus making food security a significant anxiety for most countries. The traditional methods were no longer sufficient which paved the way for technological ascents such as a substantial rise in Artificial Intelligence (AI), Internet of Things (IoT), as well as Robotics that provides high productivity, functional efficiency, flexibility, cost-effectiveness in the domain of agriculture. AI, IoT, and Robotics-based devices and methods have produced new paradigms and opportunities in agriculture. AI's existing approaches are soil management, crop diseases identification, weed identification, and management in collaboration with IoT devices. IoT has utilized automatic agricultural operations and real-time monitoring with few personnel employed in real time. The major existing applications of agricultural robotics are for the function of soil preparation, planting, monitoring, harvesting, and storage. In this paper, researchers have explored a comprehensive overview of recent implementation, scopes, opportunities, challenges, limitations, and future research instructions of AI, IoT, and Robotics based methodology in the agriculture sector.

## 1. Introduction

Farming has played a pivotal role since the inception of humankind. The farming revolution has been driving by the mechanized hands from industrial revolution to advent of AI, IoT, and Robotics technology. Magnificent changes have been rising in problem-solving, situation monitoring, decision making, market demand analysis, autopilot machinery which eventually lead the robotics approach to utilize hardware, algorithm, software, AI, ML, DL modeling, and simulation, current knowledge, and experience in the farming sector [1]. Scientists has prophesied that the earth will need 25% to 70% more food to face the food shortage demand by 2050 [2]. AI, IoT, Robotics based technology is most likely to succeed as the existing technology to power-up against upcoming challenges to successfully handle all situations.

Scientists have iterated that AI is the precursor of the current age [3]. AI has started to play the main role in daily life thus extending the way of thinking. Artificial intelligence that has been inspired by the human brain and artificial neurons is now able to compute faster than that of the human brain [4].

Today's IoT is one of the most dependable terms in the technological world [5]. The fast progress in space technology based science and data communication pertaining to that of satellite imagery also adheres to supply continuous insights pertaining to and on weather data, crop production data to the farmers for sustainable farming and development [6].

Robotics has opened a new dimension, new ground, new operating environment, new method, through a new level of performance [7]. The implementation of precision farming and actual precision management are some of the possible reactions to this prospect, which depends not only on the sensor technology but that is possible through the accurate utilization of agricultural robots. [8].

AI, IoT, Robotics technology will save countless hours of labor, collect crops in actual time, provide more food with fewer resources and reduce human involvement, ensure food security and profit for the economy, decrease the use of chemicals, minimize environmental pollution.

This chapter provides a comprehensive overview on AI, IoT, and Robotics based advanced farming using different case studies of various AI models, sensors technology, multitasking robotics. More particularly, we briefly discuss recent implementation, scopes, opportunities, challenges, limitations, and future research directions of AI, IoT, and Robotics based methodology in the farming sector.

The main contribution of our presented chapter are listed below:

● We have presented an analytical overview of AI, IoT, and Robotics based advanced farming systems.

● We have proposed the figure-based conceptual framework of AI, IoT, and Robotics in farming.

● We have outlined a comprehensive tabulated overview of problems and solutions addressed through AI, IoT, and Robotics in the Farming as well as agricultural domain.

● We have explored extensively to identify AI, IoT, and Robotics-based technological security issues and associated trust of adoption in the farming sector.

● We have addressed the challenges in the adoption of technology and have put forward a plausible futuristic conversation to move forward with adoption of technology .

## 2. Technology in Advanced Farming

Farming has foreseen a tremendous emergence in the field of AI, IoT and Robotics. In this section, we have subdivided our focal point of Artificial intelligence, Internet of Things and Robotics to provide a comprehensive overview of the topics to ascertain various implantation and scopes of these fields through its emergence. Figure 1 provides a comprehensive conceptual imagery of how technology is implemented in the farming sphere.

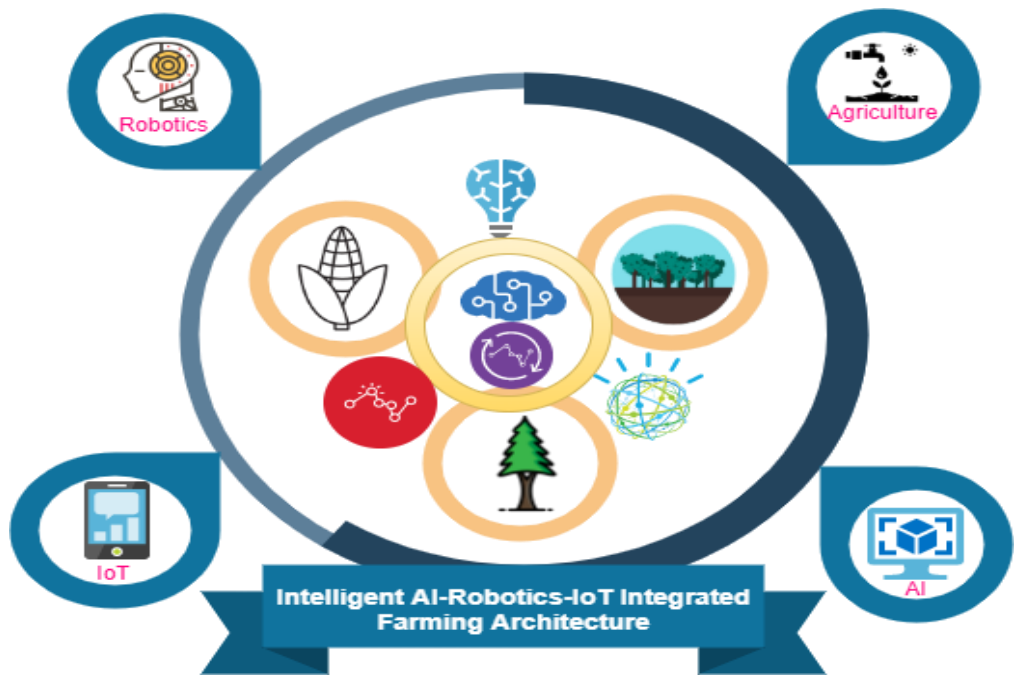

Figure 1: Conceptual Architecture of AI-Robotics-IoT Integrated Farming

2.1 AI in Advanced farming

Pareek and his team [71] have built an ANN-based seed distribution model. The presented model consists of a plate seed device, laptop, several types of motors and sensors, Arduino, liquid crystal display used for motor speed control. They applied three input neurons, five hidden neurons, one output neuron-based ANN for the cell fil prediction purpose, and PSO (Particle Swarm Optimization) employed for optimum operation purposes. Chen et.aI [72] has done transfer learning-based tomato leaves disease and pest detection methods. The authors used the PlantVillage dataset and their own collected dataset for the ResNet-18 training purposes. The transfer learning model achieved 98.06% detection accuracy. Sarijaloo and his colleagues [73] presented the corn yield performance prediction approach. For the training purpose, they used the Crop Challenge 2020 Syngenta dataset. The authors applied traditional machine learning algorithms such as TR (Random Forest), DT(Decision Tree), XGBoost, AB (Adaptive Boosting), GBM, NN. XGBoost algorithms perform best against other machine learning algorithms. Wei et.aI [74] proposed an intelligent-based fish feeding system. This system consisted of the sensor, motor, feeder, camera, synchronous wheel and belt, drive linked plate, slide, positioning platform, track, pond. Finally, the authors proved their intelligent feeding system performs better against traditional feeding systems. Wang and his colleagues [75] presented the sugar prediction approach from the sugarcane yield. Authors applied traditional machine learning algorithms like SVM, RF, LR, DT, NB, and MLP for classification purposes. The multilayer perceptron algorithm achieved the best performance than other classification algorithms.

Shadrin and his team [9] proposed a generic low-power embedded method based on AI. They ensured analysis and prediction of the dwarf tomato leaves' growth dynamics monitoring in real time with the aid of Recurrent Neural Networks (RNN). Furthermore, researchers in [10] developed a power savvy smart systematic approach of sensing coupled with AI for seed germination detection. They collected data from 18 containers totaling twenty four hundred seed images and around three thousand background images. The proposed system has used custom CNN architecture which achieved 97% accuracy at the validation set and achieved 83% average intersection over union at the testing set which is quite commendable. Sabanci [11] presented an artificial intelligence-based wheat grains detection system where the author collected 300 images. The proposed method used an artificial neural network and extreme learning machine algorithm optimized with the Artificial bee colony algorithm that achieved hundred percent classification accuracy, and ELM achieved a classification accuracy of 90%. Partel and his colleagues[12] have presented an autonomous and artificial intelligent-based herbicide sprayer system. Their sprayer system consists of three video cameras, individual nozzle control, a speed sensor, a pump, a Real-Time Kinematic GPS, relay boards, tubes, pressurized folds, etc. They have experimented with YOLOv3 with Darknet53 and Faster R-CNN with Resnet 50, Faster R-CNN with Resnet101 for real-time weed detection purposes. Faster R-CNN coupled with Resnet 50 provided the best performance in this experiment.

Similar works are done in[13] where the authors developed a precision sprayer system with artificial intelligence. They evaluated the sprayer system with artificial plants and real plants. Seyedzadeh et.al.[14] presented an artificial intelligence-based estimate of discharge of drip tape irrigation systems. The authors simulated their methodology's wide range of temperature (13−53 °C) and operating pressures (0–240 kPa) variations. They used six ANN architectures, neuro-fuzzy sub-clustering, neuro-fuzzy c-Means clustering, least-square support vector machine with radial basis function kernel, and least-square support vector machine with linear kernel for operating pressure and temperature measurement purposes. Overall, the authors noted that the least-square SVM along with RBF kernel, neuro-fuzzy sub-clustering, ANN with Levenberg-Marquardt achieved highest performance accuracy. Kaab and his team [15] proposed artificial intelligence-based environment based adverse and positive impacts of sugarcane production measured prediction systems. The authors used artificial neural networks and an adaptive neuro-fuzzy inference system (ANFIS) where ANFIS performed better than ANN. Furthermore, researchers in [16] presented deep residual CNN-based invertebrate pests detection methods. The author's dataset included more than six thousand plus images. The dataset was constructed as field images, virtual images, random rotation, random crop, random resize, and random-contrast adjustment[16] and tested on standard pre-trained networks. The authors in [17] proposed two AI models, SVM and ANN for maize evapotranspiration estimation. The authors collected input data from maize fields. The input data consisted of meteorological variables, leaf area index, and plant heights, continuous measurements of evapotranspiration. The genetic artificial neural network model architecture performed better than the SVM models. Picon et.al. [18] presented three different CNN models for plant disease classification. The authors collected one-hundred-thousand real-field crop images taken by cell phones. They also used ResNet50 baseline architecture. The best model achieved 0.98 classification accuracy.

A conceptual architecture of AI in farming can be overviewed in Figure 2 and a summarized version can be visualized in Table 1 for understanding various problems in agricultural domain that are being addressed via deep learning and machine learning models.

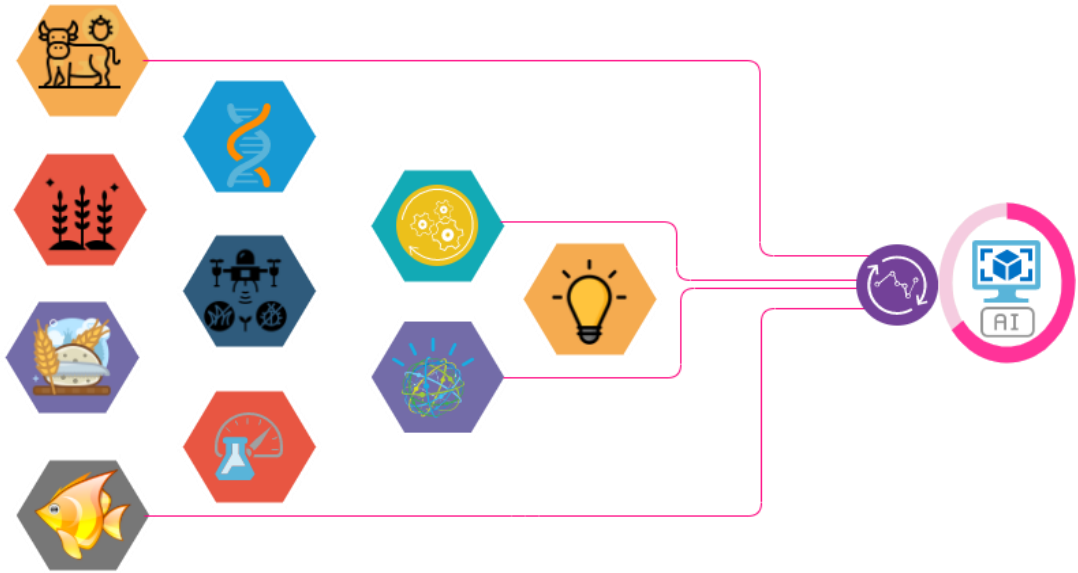

**Figure 2: A Conceptual Architecture of AI in Farming**

**Table 1: A Comprehensive overview of problems and solutions addressed through DL/ML models in the Farming domain.**

| Article | Addressed Problem | DL/ML Model | Accuracy | Dataset |
|---|---|---|---|---|
| [19] | Multi classe cassava leaf disease detection | CNN | 93% | Kaggle |
| [20] | Multi classe apple leaf disease detection | Custom CNN, InceptionV3, VGG16 | 99.6%, 99.4%, 99.5% | PlantVillage |
| [21] | Multi classes tea leaves disease detection | SVM, DT, RF, C-DCGAN + VGG16 | 70%, 71.6%, 75% 90% | Dataset was curated by authors |
| [22] | Citrus aurantium diseases detection | Conditional adversarial autoencoders | 53.4% | CUB, AWA2, APY |
| [23] | Multi classes grape leaf disease classification | GAN | 92.44% | PlantVillage |

| [24] | Plant seedlings classification | VGG16 | 99.48% | Aarhus University, University of Southern Denmark |
|---|---|---|---|---|
| [25] | Plant disease detection | AlexNet, AlexNetOWTBn, GoogLeNet, Overfeat, VGG | 99.06%, 99.49%, 97.27%, 98.96%, 99.53% | PlantVillage |
| [26] | Sugar beet leaf spot disease detection | Faster R-CNN | 95.48% | Dataset was curated by authors |
| [27] | Multi classes potato disease classification | VGG | 83% - 96% | Dataset was curated by authors |
| [28] | Yellow rust disease detection | Inception-Resnet, Random forest classifier | 85%, 77% | Dataset was curated by authors |
| [29] | Plant Disease Detection | DL | 93.67% | PlantVillage |

## 2.2 IoT in Advanced farming

Podder and his team [76] developed an IoT system that decides irrigation operations begin or end based on weather and farming land conditions. They also provide the farm remote controlling system. This IoT system consisted of various sensors like moisture, freeze, wind, humidity, pH, etc. The ESP8266 tools are used for system control purposes. Wi-Fi module used for data transmission purposes. Khan et.aI [77] presented IoT-based farming monitoring methods for Onion Farms bolting reduction. The authors divided their system into three levels. Level one used various sensors like temperature, humidity, and light intensity to collect environmental parameters. Level two, ThingSpeak and Wi-Fi employed for data transmission purposes. Level three, data visualization and monitoring system. Almalki and his colleagues [78] have proposed IoT and Unmanned Aerial vehicle-based farming monitoring approaches. They employed various types of sensors like temperature, rain, humidity, soil moisture, solar radiation to collect environmental parameters. The drone system consists of a camera, sensor, motors, Raspberry Pi 3 microcontroller unit, propellers, flight controller, battery, and long-range wide-area transmission system. Cloud computing layer applied to the process, share, visualization, operations, decision-making purposes. Addo-Tenkorang and his team [30] presented a low-cost animal monitoring system. Authors included their system with animal health history, location tracking, birth record, and ownership history. The proposed system hardware consists of a 3D Printer, Micro Tempered Solar Panels, Micro Solar Inverters, Micro Rechargeable cells, RFID tags, GPRS, SIM Cards with roaming function. Software components Cloud Database System, Web Services Interfaces, Mobile Gadgets, Operating system, Google Earth with proper encryption and security protocol in place. The researchers in [31] developed a water monitoring system. The authors used sensors to calculate Water Temperature, pH, TDS, Turbidity, and Conductivity. Sensors were connected to the Arduino UNO. The Wi-Fi module was used for data transmission purposes. The IoT-based API was used to view sensor data. Authors in [32] proposed a farm monitoring system with the main aim to detect intensity emitted by light, moisture measurement of the soil, and temperature reading of the surroundings. The sensors are used to measure each parameter. These sensors sent data to the Arduino followed by processed data to Raspberry PI where they used MIT AppInventor2 android application, IoT-based gateway to web server monitoring system for designing the system. Bodake et.al. [33] developed a soil-based fertilizer recommendation system which has a similar output to that of [32]. Similarly, authors in[34] utilized an automated farming system for orchids cultivation. The authors used sensors to collect parameters pertaining to the environment such as temperature, humidity, light, and soil moisture [34]. Righi and his team [35] presented a dairy cattle management system. The authors divided their approach into two parts : (i) MooField and (ii) MooServer.The first part acted as the data collection service as well as service used in the field whereas the second system is used to store, analyze and generate insight from the data. Their model achieved 94,3% accuracy for prediction purposes. Chung et.al. [36] developed a solar-powered real-time crop growth monitoring system. Temperature, humidity, soil moisture, light, chlorophyll, and CO2 monitoring each sensor were used for input purposes. ZigBee-based wireless network was utilized for data transmission purposes. The sensing data was applied for evaluation of the environmental condition, risk measurement and management, and data-driven decision-making. The authors in [37] presented a secure remote user authentication with agriculture field monitoring system. In [37], the sensors collected environmental variables such as temperature, humidity, light, soil pH, soil moisture, Co2. Similar types of monitoring systems were developed in[38] where the authors proposed Brassica Chinensis growth monitoring and indoor climate control system via the light-emitting diode. The authors have experimented with four different light treatments. They analyzed plant growth based on leaf count, height, dry weight, and chlorophyll. The authors in [39] presented an IoT based irrigation approach and real-time monitoring system. The proposed method has been divided into two-parts; one of which in the control system, they used submersible motors, relay modules such as 5V10A2 to control the motors, overhead tanks, excess water collection tanks, ultrasonic sensors to control the water level in tanks, moisture sensors to control the soil moisture. All sensors have interacted with Raspberry PI 3. On the other hand, Farmer's cockpit has a real-time monitoring application provided by Apple's integration. The weather forecast authors used yahoo weather APIs.

A conceptual architecture of IoT in farming can be overviewed in Figure 3 and a summarized version can be visualized in Table 2 for understanding various problems in the agricultural domain that are being addressed via IoT induced devices.

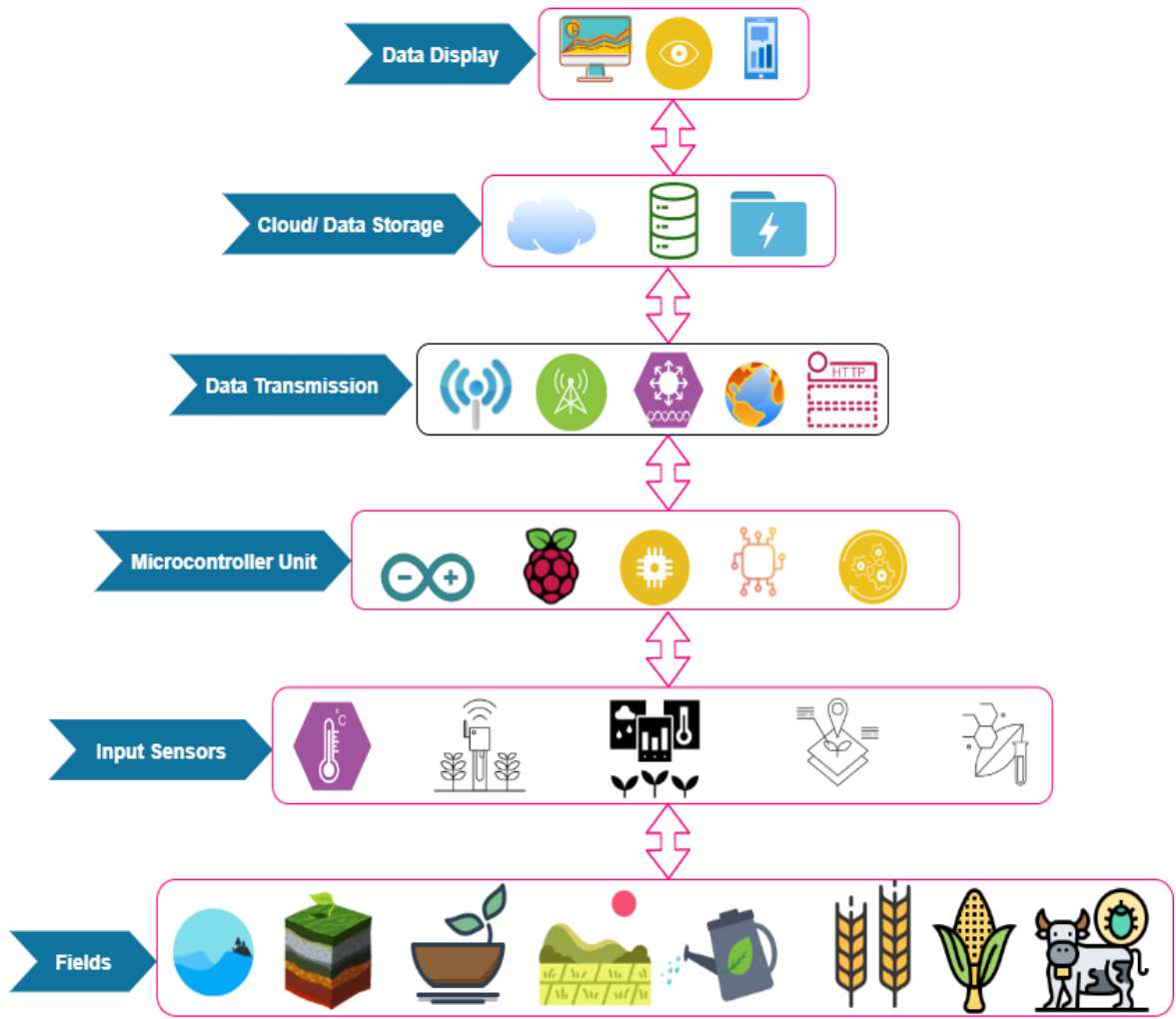

**Figure 3: A Conceptual architecture of IoT in Intelligent Farming**

**Table 2: A Comprehensive table with problem overview and related IoT devices for plausible solution approaches in Intelligent farming**

| Article | Addressed Problem | Input Parameter | Data Controller | Data Transmission | Data Display |
|---|---|---|---|---|---|
| [40] | Mushroom cultivation monitoring system | Humidity sensor MQ135 CO2 gas sensors | Arduino IDE | Wi-Fi | ThingSpeak |
| [41] | Potato late blight prevention method | Temperature, Humidity, Freeze sensors | Cloud | Wifi, Bluetooth, ZigBee, etc | Smartphone, Laptop, Desktop |
| [42] | Agriculture monitoring system | Temperature, Humidity sensors | Arduino UNO | Wi-Fi, Ethernet Shield | Smartphone, Laptop, Desktop |
| [43] | Agriculture monitoring system in rural areas. | Temperature, Humidity, Pressure, Luminosity, Soil Moisture etc sensors | RaspberryPi | Wi-Fi | Web-based interface |
| [44] | Watering management System | Water level, Soil moisture, Temperature, Humidity, and Rain intensity sensors | Arduino | Wi-Fi | LCD screen |
| [45] | Simulated water management | Various simulated sensors | Various simulated engine | Wi-Fi | Application Services |

| [46] | Strawberry disease prediction System | EC1, pH, Temperature, Humidity, CO2 sensors | Cloud | Long-range wide area network | Smartphone |
|---|---|---|---|---|---|
| [47] | Cow health monitoring system | Accelerometer magnetometer sensors | Raspberry Pi | Wi-Fi | Android Application |
| [48] | Agricultural monitoring system for early plant disease forecast | Air Temperature, Air Humidity, Soil Temperature, Soil Volume Water Content, Electrical Conductivity, Wind Speed, Wind Direction, Rain Meter, Solar Radiation, Leaf Wetness sensors | Arduino ADCs | Cellular network | Web application |
| [49] | Potato disease prevention system | Humidity and Temperature sensors | NodeMCU | Wifi, Bluetooth, ZigBee and 3G/GPRS | Web application |

**2.3 Robotics in Advanced Farming**

Rajalakshmi et.aI [79] developed a vision-based robot that performs various farming operations like seed sowing, spraying, and watering. This robotics system consists of a camera, RaspberryPi, Arduino, power supply unit, pump, etc. For training purposes, the authors used the Plant-Village dataset. YOLO model used for weed detection purposes. Khadatkar and his team [80] presented an automatic transplanting system. The proposed device consists of the stepper motor, DC motor, 12 V power supply battery, feed roller unit, the pro-tray belt, two L-shaped rotating fingers, ground wheel, and Arduino Uno microcontroller unit. The laptop is employed to control the whole system.

Quaglia and his colleagues [50] developed an Unmanned Ground Vehicle(UGV) robotic system to collect crop samples and collaborate with the drone. This robotic system consists of seven actuators, a robotic arm, sensors, control devices. Equipped with solar panels, the authors provide two operative modalities which are remote monitoring and an on-board analytic tool. Jawad and his team [51] have constructed a wireless power transfer-based efficient drone battery charging system. This drone can collect climate conditions information from the agriculture field. They measured drone battery performance based on two strategies: sleep and active. The authors successfully extended 851 minutes of battery life, achieving 96.9% battery power saving. Similar research work of curating an autonomous vehicle and a spraying system have been done in [52]. Furthermore, Pulido Fentanes et.aI [53] developed an autonomous mobile robot system for creating high-quality soil moisture maps. The cosmic-ray sensor was applied to count fast neutrons. The authors evaluated their method using soil moisture data. For making better quality soil moisture patterns, they applied adaptive measurement intervals and adaptive sampling strategies. Kim and his colleagues [54] have done an autonomously machine-vision-based crop height measurement system. The auto-exposure mode was applied to capture images, OpenCV library has been used to process images. The presented system achieved less than 5% error for five different crops. Kumar and Ashok [55] have developed an automated seed sowing robot. The robot consisted of the actuator, sensing device, stepper motors, seed handling unit, microprocessor, servo motors, power transmission, communication, data controlling processing unit. The author's claimed that seed sowing robots reduces time and human involvement in the agriculture sector. Thorp and his colleagues [56] have created a drone-based daily soil water balance measured model. Pix4D software was employed to process images. Processed images utilized georeferencing purposes. For visualization and analysis of the data, they used ENVI software. Mahmud and his team [57] have done the spraying robot path planning system. The authors constructed a mobile sprayer robot. They designed a greenhouse environment and a virtual environment where they tried to decipher the shortest path and lowest routing angle. Overall Non-dominated Sorting Genetic Algorithm III achieved better performance. Distributed multi quadrotor UAV-based crop monitoring systems have been done in [58]. The author created a virtual environment that used Gazebo robot simulator framework. The presented system simulated three Iris model quadrotor UAVs. Similar to Mahmud in [57], Srivastava and his colleagues [59] did an optimal route algorithm for UAV's based fertilizers and pesticide spray in the field where Traveling salesman algorithm was applied to find the shortest path. Raeva et.aI [60] did the corn and barley map monitoring system based on UAVs. They used multispectral and thermal sensors. The multiSPEC 4C and senseFly thermoMAP cameras are employed to capture image information from the fields. The UAVs system captured all information from March to August 2016. Pix4mapper was used for image processing purposes.

A conceptual architecture of robotics in farming can be overviewed in Figure 4 and a summarized version can be visualized in Table 3 for understanding various problems in agricultural domain that are being addressed via robotic devices.

**Table 3: A comprehensive tabulated overview of Robotics in Advanced Farming**

| Article | Addressed Problem | Platform | Components | Controlling System |
|---|---|---|---|---|
| [61] | Fish monitoring system | Fish robotics | ABS plastic, SMA wires, SMA drivers, current, thick polycarbonate backbone, temperature sensors, synthetic skin | PID controller |
| [62] | Tree trunk detection method | Mobile robot | Ultrasonic sensors, camera, angle sensor, stepper motor, four-wheels, electronic compass, etc. | PI controller |
| [63] | Automatic spraying system | UGVs robot | Nozzle chamber, Kinect camera, liquid pump and tank, Arduino, Two electromotors, etc | Laptop |
| [64] | Autonomous navigation algorithm for wolfberry orchards | UGVs robot | Camera, crawler, steering cylinder, steering clutch, gasoline engine, wire rope, air pump, air tank, digital display pressure gauge, oil mist device, | Fuzzy controller |

| | | | etc. | |
|---|---|---|---|---|
| [65] | ROS emulation toolkit | UGVs robotics | GPS, RTK-GNSS, LiDAR, camera, ultrasonic sensors, four-wheel-driving four-wheel-steering electric vehicles, etc | Computer |
| [66] | Simulated plants monitoring system | Mobile robot | Sensors, routing algorithm, adaptive search algorithm etc. | Simulated Controlling System |
| [67] | Tomato harvesting method | Mobile robot | Camera, ultrasonic sensor, raspberry pi, arduino Mega 2560, servo motors, arm robot, power system, etc. | Computer |
| [68] | Robot mower track and performance measurement system | Rover | RTK-GNSS, Wi-Fi and Bluetooth, cables, power system, Intel® Edison module, Qprel®srl etc. | Computer |
| [69] | Robotics based modeling and control methods | Mobile robot | Drive wheeled vehicles, arduino UNO, DC motors, IMU sensor, tracks device, mobile tank robot kit 10022 etc. | PID, Fuzzy logic |
| [70] | Real-time soil electrical resistivity measurement system | UGVs robot | Four-wheel, four-probes wenner, DC motors, DGPS, stainless steel, fiber isolation rings, cable, 24-volt battery, inject electric current, software, etc. | Computer |

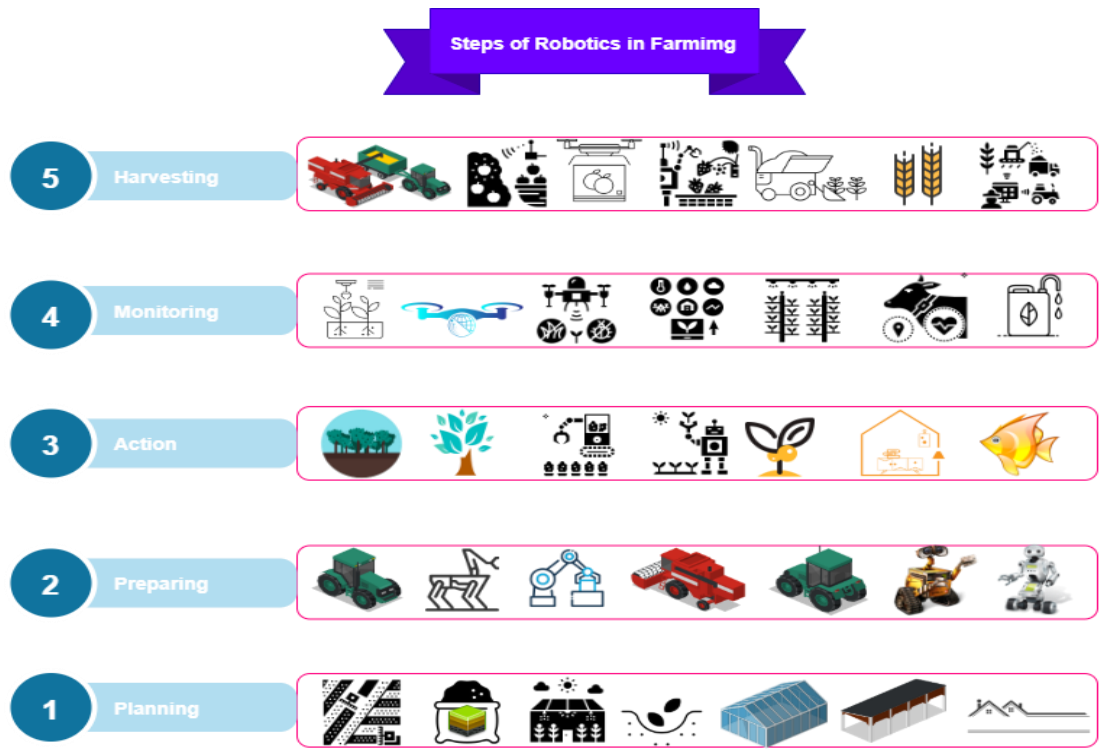

**Figure 4: A Conceptual Framework of Robotics in Farming**

## 3. Challenges in Adoption of Technology

There are obvious challenges and factors that are in effect for the acceptance and adoption of the technology. The challenges can be categorized into various stakeholders who would be using technology for advanced farming and the perceived issues that will be raised here along with security concerns that hinders the adoption of technology in agriculture. There are perceived issues from human angel as well as issues related to research and development that are needed to address to improve the user acceptance of the technology.

The main stakeholder for the technology is the farmers who would be using the technology to improve the farming sector. However, there is a big disparity in technical knowledge among the farmer in all over the world. Although, farmers in north America have successfully indulged themselves in technology to create a best-case working scenario, the situation is fully different in Asia and Africa where technical perspectives are overlooked in general. The main reason for this gap is, lack of resource personnel who would create a knowledge sharing platform as well as the lack of interest of the farmers to accept the technology as new normal in their life. The main reason of lack of interest is the lack of overall awareness and knowledge that how exactly technology may provide tangible development monetarily as well as in an eco-friendly way to create a sustainable development architecture for all stakeholders. As for teaching technology in root level, although various research works are carried out in research field, the practicality and applicability of these research works are often overlooked. Apart from applicability, there is a substantial gap in field work where trained professional can go around teaching the importance of technology which are the embargo towards accepting the technology.

There are minor and major security issues related to hacking in cases of AI, IoT and Robotics which needs further attentions from the researcher communities. The main problem due to security issues would lead to disaster in crop harvesting which would lead to food shortage thus raising an issue within the concerned stakeholders to be reluctant to adopt the technology. Furthermore, challenges, and scopes are continuous process that through check and balance needs to be addressed and amended to provide a sustainable development with tangible outcome for the whole community.

## 4. Conclusion

This research work has pointed out various developments that has been carried out in the field of Artificial Intelligence, Internet of Things and Robotics with a short overview of challenges that may be faced while adopting such technology. The research work provided a comprehensive review of aforementioned sectors of technologies which provided an idea on current status of intelligent farming. Furthermore, the research work also pointed out the perceived challenges as well as the security concerns that are needed to be addressed from the standpoint of human computer interaction design, usability studies and strengthening the securities. The authors believe that this review article would point the researchers working in agriculture domain towards current status and future possibilities thus contributing to development of intelligent technology-based farming and agriculture research.